\documentclass[conference]{IEEEtran}
\IEEEoverridecommandlockouts
\usepackage{cite}
\usepackage[colorlinks=true, linkcolor=black, citecolor=black, urlcolor=blue]{hyperref}
\usepackage{amsmath,amssymb,amsfonts}
\usepackage{graphicx}
\usepackage{textcomp}
\usepackage{xcolor}
\usepackage{tabularx}
\usepackage{cite}
\usepackage{array}
\usepackage{multicol}
\usepackage{multirow}
\usepackage{makecell}
\usepackage{booktabs} 
\usepackage{algorithmic}
\usepackage[linesnumbered,ruled,vlined]{algorithm2e}
\usepackage{csquotes}

\def\BibTeX{{\rm B\kern-.05em{\sc i\kern-.025em b}\kern-.08em
    T\kern-.1667em\lower.7ex\hbox{E}\kern-.125emX}}
\begin{document}

\title{LLM-empowered Agents Simulation Framework for Scenario Generation in Service Ecosystem Governance
\\
\thanks{* Corresponding author: Lizhen Cui and Xiao Xue; e-mail: clz@sdu.edu.cn; jzxuexiao@tju.edu.cn}
}

\author{\IEEEauthorblockN{
Deyu Zhou\textsuperscript{1,2},  
Yuqi Hou\textsuperscript{1},  
Xiao Xue\textsuperscript{3,4,5*}, 
Xudong Lu\textsuperscript{1,2}, 
Qingzhong Li\textsuperscript{1,6}, 
Lizhen Cui\textsuperscript{1,2*}} \\

\IEEEauthorblockA{
\textit{\textsuperscript{1} School of Software, Shandong University, Jinan, China} \\  
\textit{\textsuperscript{2} Joint SDU-NTU Centre for Artificial Intelligence Research (C-FAIR), Shandong University, Jinan, China} \\ 
\textit{\textsuperscript{3} College of Intelligence and Computing, Tianjin University, Tianjin, China} \\
\textit{\textit{\textsuperscript{4} Tianjin Key Laboratory of Healthy Habitat and Smart Technology, Tianjin, China }} \\ 
\textit{\textit{\textsuperscript{5} Laboratory of Computation and Analytics of Complex Management Systems, Tianjin University, Tianjin, China}} \\
\textit{\textsuperscript{6} ShanDa Dareway Software Co. Ltd, Jinan, China } \\
} \\
Email: zhoudeyu@mail.sdu.edu.cn, 202100300341@mail.sdu.edu.cn, jzxuexiao@tju.edu.cn, dongxul@sdu.edu.cn, \\ lqz@sdu.edu.cn, clz@sdu.edu.cn}

\maketitle

\begin{abstract}
As the social environment is growing more complex and collaboration is deepening, factors affecting the healthy development of service ecosystem are constantly changing and diverse, making its governance a crucial research issue. Applying the scenario analysis method and conducting scenario rehearsals by constructing an experimental system before managers make decisions, losses caused by wrong decisions can be largely avoided. However, it relies on predefined rules to construct scenarios and faces challenges such as limited information, a large number of influencing factors, and the difficulty of measuring social elements. These challenges limit the quality and efficiency of generating social and uncertain scenarios for the service ecosystem. Therefore, we propose a scenario generator design method, which adaptively coordinates three Large Language Model (LLM) empowered agents that autonomously optimize experimental schemes to construct an experimental system and generate high quality scenarios. Specifically, the Environment Agent (EA) generates social environment including extremes, the Social Agent (SA) generates social collaboration structure, and the Planner Agent (PA) couples task-role relationships and plans task solutions. These agents work in coordination, with the PA adjusting the experimental scheme in real time by perceiving the states of each agent and these generating scenarios. Experiments on the ProgrammableWeb dataset illustrate our method generates more accurate scenarios more efficiently, and innovatively provides an effective way for service ecosystem governance related experimental system construction.
\end{abstract}


\section{Introduction}
As the social landscape is redefined under the service logic, the service system has been gradually transcending the boundaries of traditional industries and evolving into a service ecosystem that is collaboratively operated through the division of labor among diverse intelligent service entities, including individuals, communities, intelligent robots, and AI platforms. 

\begin{figure}[htbp]
\centering
\includegraphics[width=0.48\textwidth]{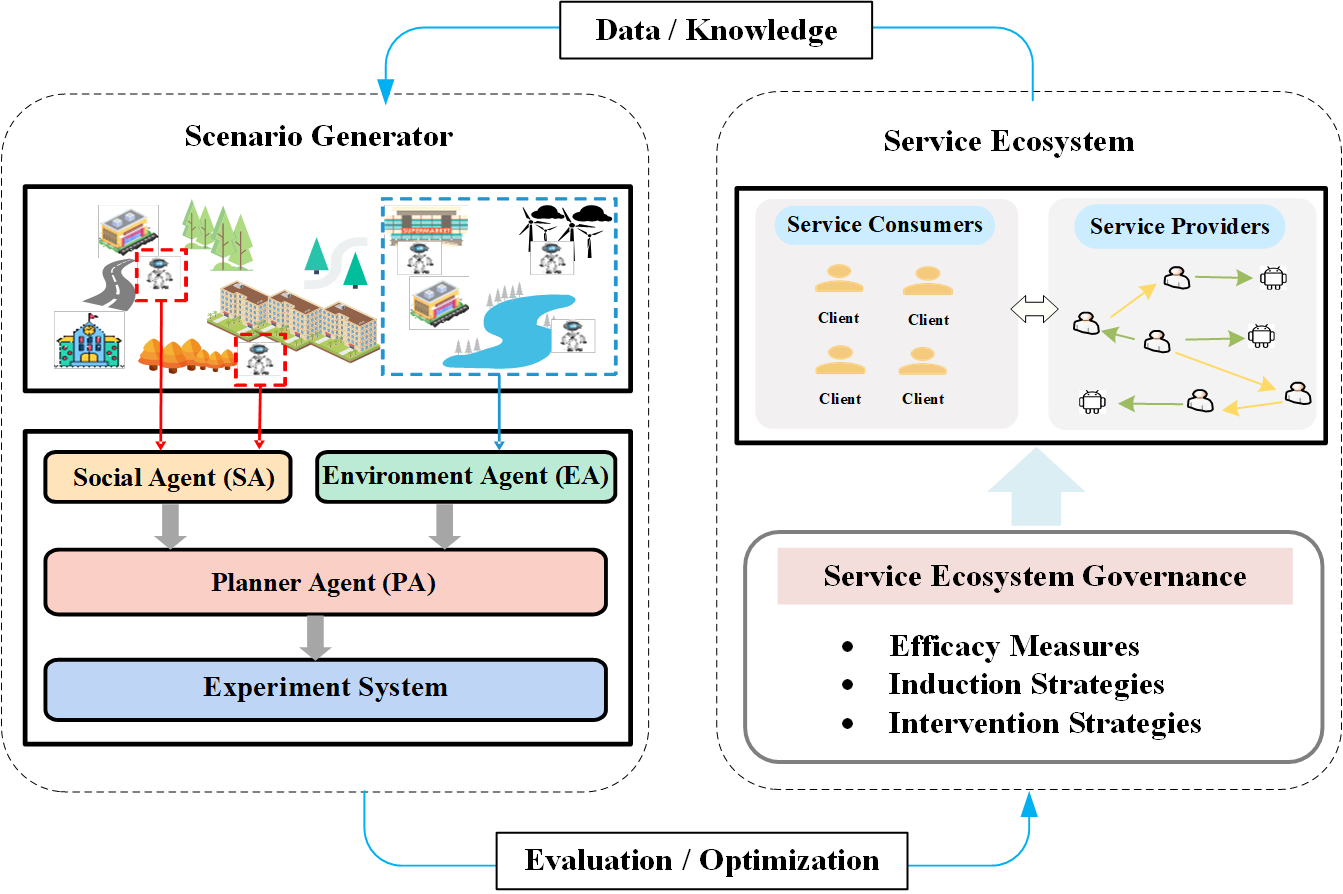}
\caption{Relationship between scenario generator and service ecosystem. Based on the data and knowedge from service ecosystem, scenario generator collaboratively generates experimental schemes by  PA, SA and EA. Then, the experiment system produces a series of scenarios to assist in system governance and optimization. }
\label{fig1}
\end{figure}

In this novel service system, various entities, namely service providers, service consumers, form a dynamic collaboration network via resource virtualization and servitization, thereby enabling cross-industry value co-creation (the right part of Fig.\ref{fig1}). Its operation mechanism is characterized by three prominent features. Firstly, social demands exhibit a dual nature of dynamism and complexity. The evolution of consumer preferences and market fluctuations interactively propel the rapid reconstruction of service scenarios. Secondly, there exists a significant heterogeneity in the intelligence of entities. The emotional cognition of humans and the rational decision-making of AI give rise to a multi-dimensional interaction pattern. Thirdly, the system connections yield non-linear effects, where a strategic adjustment of a single node has the potential to trigger a global cascading reaction \cite{7078890}. These features render the service ecosystem a typical complex adaptive system, posing severe challenges to traditional governance methods when confronted with scenarios such as abrupt changes in demand and the failure of entity collaboration \cite{xue2023computational}.

Currently, research on the governance of service ecosystems predominantly employs three methodologies: data-driven modeling and anomaly monitoring, macro-evolutionary analysis of complex networks, and experimental simulation based on Agent-based Modeling (ABM) \cite{xue2022research}. Despite the progress achieved by these methods in specific scenarios, they share three common limitations. 

\begin{itemize}
    \item The construction of the experimental system relies on predefined rules, making it arduous to model the heterogeneity of entity intelligence and the dynamics of social relationships.
    \item There is an insufficiency in the generation of extreme scenarios. Existing data-driven methods are constrained by the distribution of historical data, resulting in limited coverage of \enquote{black swan} events. 
    \item The complexity of the system means that parameter calibration requires a large amount of human resources and time, making it difficult to adapt to the real-time evolution of the system promptly. 
\end{itemize}

These issues have led to deficiencies such as scenario distortion and decision-making delays in existing governance strategies, necessitating the exploration of a new methodological framework. Therefore, this paper proposes a scenario generator design method driven by LLM-empowered agents, a core component of the experimental system. The framework integrates three collaborative agents (i.e., the left part of Fig.\ref{fig1}): \textbf{Planner Agent (PA)} dynamically compiles domain rules via RAG and program-aided prompting, translating natural language constraints into executable logic. \textbf{Environment Agent (EA)} employs LLM-based semantic deconstruction and adversarial prompt engineering to generate extreme scenarios (e.g., policy shifts, demand surges), overcoming traditional statistical constraints. \textbf{Social Agent (SA)} models heterogeneous entities' behavioral strategies by distilling social relationship networks through LLM-driven cognitive simulations.

The scenario generator synthesizes scenarios by interacting with the data knowledge base from service ecosystem\cite{xue2024computational2}. These scenarios undergo experimental deduction, yielding decision insights to optimize the service ecosystem’s matching logic and operator strategies. The closed-loop \enquote{generate-validate-optimize} mechanism enables adaptive governance of complex service ecosystems under uncertainty. Experiments demonstrate significant improvements in scenario generation efficiency and feature coverage, offering a novel tool for digital-era service ecosystem governance \cite{lu2021computational}.

The academic contributions of this paper are manifested in the following aspects: 
\begin{itemize}
    \item Establishing a scenario generation framework based on the collaboration of multiple agents of the LLM, breaking through the paradigm of predefined rules through semantic-driven approaches.
    \item Proposing a backbone extraction method for the collaborative relation of heterogeneous agents, enabling the quantitative characterization of the level of human intelligence and correlation relationships by utilizing the social cognitive simulation capabilities of the LLM.  
    \item Constructing an adversarial generation-verification closed loop for extreme scenarios, enhancing the coverage of long-tail events through the causal reasoning capabilities of the LLM.
\end{itemize}

The rest of this paper is organized as follows: Section \ref{bk} presents the research background and motivation; Section \ref{body} presents the modeling framework of the scenario generator; Section \ref{em} gives the design and analysis of computational experiments; Section \ref{cl} discusses and concludes the paper.

\section{Background and Motivation} \label{bk}
Influencing factors in the evolution process of service ecosystem are complex and difficult to confirm. They are not only affected by human free will and subjective initiative, but sometimes accidental events may also change the evolution path, posing huge challenges to the governance of service ecosystems \cite{xue2021computational}. The commonly used analysis methods currently used by researchers can be summarized into the following three types:

\textbf{1) Data-driven analysis} identifies potential system failures by monitoring operational data, detecting anomalies like outliers, fluctuation points, and abnormal time series\cite{xue2024computational}. These methods include statistical models (e.g., hypothesis testing), proximity-based techniques (e.g., KNN), and density-based approaches (e.g., LOF). Recent advances emphasize multimodal data integration: Hongyi Liu et al. proposed an adversarial contrastive learning framework to fuse metrics and logs, improving anomaly detection robustness in microservices \cite{liu2024uac}; Mingyi Liu et al. designed a multilayer network model capturing cross-layer interactions (functionality, collaboration, competition) to predict system evolution and enable multi-granularity governance \cite{liu2023data}.

\textbf{2) Complex network analysis} effectively characterizes service ecosystem structures and evolution. Shin et al. used centrality metrics (degree, betweenness) and clustering to identify research trends from 340 publications \cite{shin2021study}. Wang Z. and Tu Z. proposed a service community framework to detect historical patterns (e.g., cluster differentiation, strategy convergence) for trend prediction \cite{liu2022community}. Huang Keman et al. applied this approach to web service ecosystems, quantifying static topologies (node connectivity) and dynamic behaviors (service lifecycles)\cite{huang2012empirical}.  Ali Yassin et al. developed netbone, a Python package evaluating backbone extraction techniques in weighted networks, covering statistical, structural, and hybrid methods\cite{yassin2023evaluation}. While powerful for macro-level analysis, these methods depend on predefined topologies and overlook micro-level interactions.

\textbf{3) Agent-based modeling} addresses this gap by simulating autonomous agents to uncover system-level evolution mechanisms. Villalba et al. emphasized self-organization, self-adaptability, and long-lasting evolvability in service systems, validating designs through multi-agent simulations \cite{villalba2011towards}. Ye et al. modeled services as autonomous agents, developing an agent-based self-evolving composition method covering five stages: discovery, selection, negotiation, execution, and adaptation \cite{ye2016agent}. Lim et al. designed the AppEco model to simulate mobile app ecosystems, evaluating development strategies’ impacts on downloads, diversity, and adoption rates \cite{lim2012successful}. Xue et al. further introduced a social learning framework to model competitive-cooperative evolution across individual, organizational, and societal layers \cite{xue2018social, zhou2022sle2}.

In summary, current research on service ecosystems predominantly focuses on the statistical analysis and replication of macro-level system characteristics, which heavily relies on researchers' a priori knowledge and assumptions about the system \cite{xiao2023putational}. Three critical challenges persist:
\begin{itemize}
     \item Generating extreme scenarios and long-tail distribution events to address \enquote{black swan} risks that traditional data-driven approaches fail to capture;
     \item Quantifying qualitative features such as the diverse intelligence levels of heterogeneous agents and complex social relationship networks, thereby transcending the limitations of predefined rules;
    \item Achieving low-cost parameter calibration and real-time evolutionary adaptation to reduce experimental costs while ensuring the dynamic optimization of complex service ecosystems.
\end{itemize}

Recent advances in LLM technology and scenario analysis methods have expanded service ecosystem governance opportunities \cite{xue2023chatgpt}. Scenario analysis, formalized by Pierre Wack in 1970s, systematically evaluates system dynamics to design plausible futures through narrative-driven frameworks. Qualitative approaches rely on expert knowledge and visual tools (e.g., diagrams) to map factor relationships and integrate insights into future projections \cite{jung2023toward,davis2023towards}. While modern tools enhance data visualization and simulation (e.g., Sora’s text-to-scene generation), their output quality depends heavily on scenario descriptions. Quantitative methods, like scenario generators, automate parameter-based systems (e.g., autonomous driving\cite{lu2023test, xue2023anew}, renewable energy systems\cite{li2020review}) by abstracting problems into models with predefined distributions. However, they struggle with complex social systems due to oversimplification and inability to capture multi-element interactions.

Relying on LLM technology, this paper applies the scenario analysis method to the research on the governance of service ecosystems and proposes the scenario generator design method for service ecosystem governance driven by LLM empowered agents simulation to address the above challenges.  

\section{Design Methodology of the Scenario Generator Framework} \label{body}

\begin{figure*}[htbp]
\centering
\includegraphics[width=0.99\textwidth]{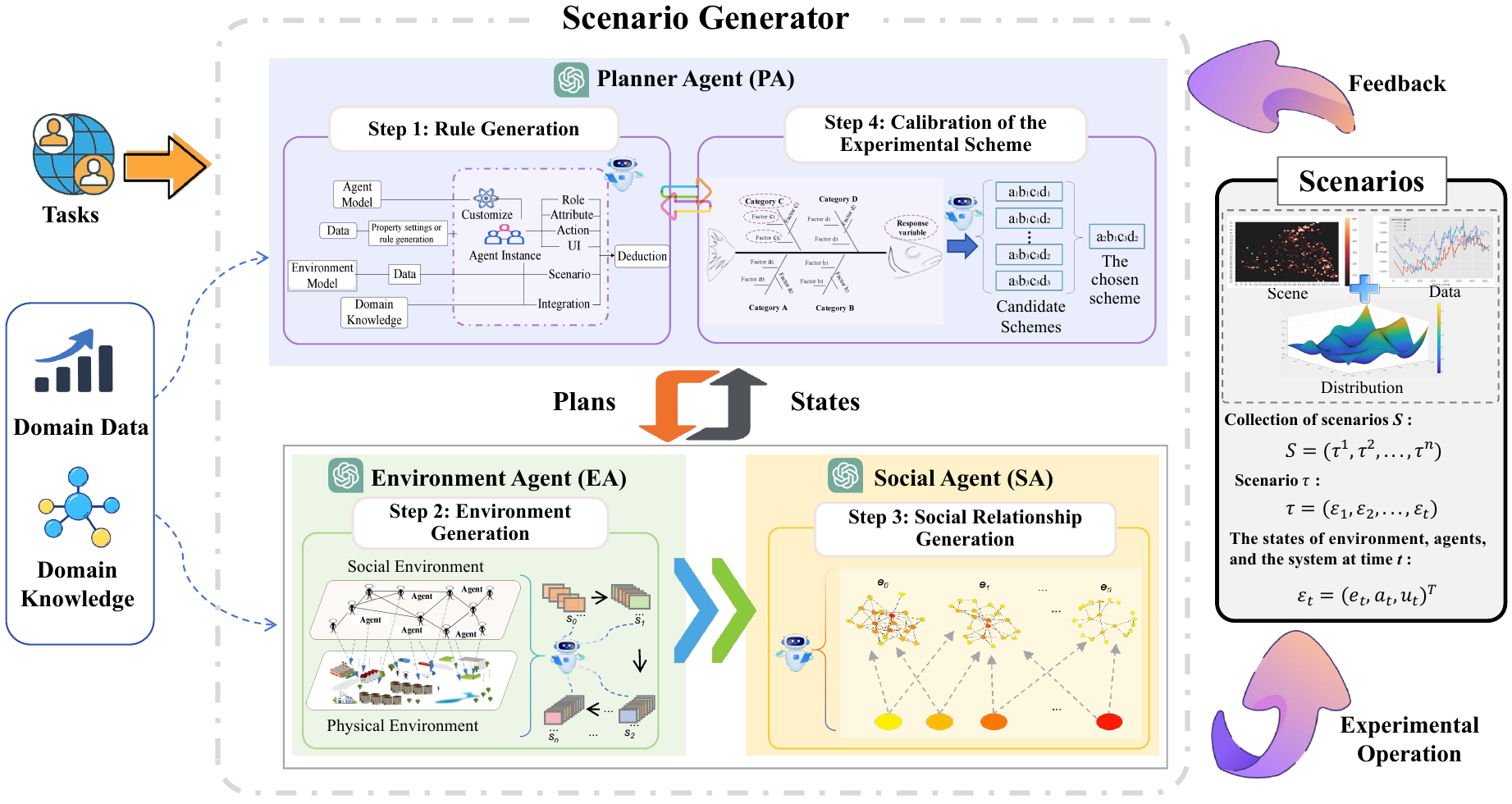}
\caption{The framework of scenario generator design method. Rules generated by PA, the environment model created by EA, and social relationships formed by SA are developed to experimental scheme. Finally, it runs to output a scenario set for experimental analysis, and the scheme is calibrated through feedback.}
\label{fig2}
\end{figure*}

This section initially expounds upon the overarching framework of the scenario generator, as shown in Fig.\ref{fig2}. Subsequently, a detailed elaboration is conducted for each module within the framework. Specifically, it involves the generation of extreme scenarios by the Environment Agent (EA), and during the generation process of the Social Agent (SA), the extraction of the backbone network pertaining to the subject-collaborative process. Ultimately, the rule - generation mechanism within the Planner Agent (PA) and the process of scheme calibration are expounded.

\subsection{Problem Formulation}
The scenario of a service ecosystem can be conceptualized as a holistic outcome derived from the interactions between the demand environment and service entities over a temporal interval $T$. This framework systematically delineates the constituent elements encompassing demand environment characteristics, operational status of service supply-side entities, interaction networks among stakeholders, along with response variables that reflect system efficacy states. 

Within the time period $T$ ($T$ is the dimension of the scenario), and on a known probability space, the parameter vector $\varepsilon_t (t = 1,2,\cdots,T)$ is composed of random variables that follow a finite discrete distribution. Then $\tau = (\varepsilon_1,\varepsilon_2,\cdots,\varepsilon_T)$ forms a matrix composed of a $T$-dimensional random vector. A realization value $\tau^k = (\varepsilon_1,\varepsilon_2,\cdots,\varepsilon_{T_k})$ of $\tau$ is called a scenario. The scenario generation method is used to generate a set $S(S=\{\tau^1,\tau^2,\cdots,\tau^n\})$ containing $n$ scenarios. Each scenario represents a situation that may occur in reality and corresponds to a scenario occurrence probability $p_s$. The larger the scenario probability, the greater the possibility of occurrence. The sum of all scenario probabilities is 1. Considering comprehensive influencing factors (including the environment and the subject) and response variables, the parameter vector $\varepsilon_t$ at time $t$ is defined as follows:
\begin{align}
\label{eq:S}
\varepsilon_t(e_t, a_t, u_t) &=(E(e_t), I(a_t), U(u_t))^T,\\
\tau^c &=\min\{G_{Scenes}(S)|e, a, u\},\\
X &\to S, \forall \tau^k \in S, \tau^c \subset S
\end{align}
Here, $E()$ represents the method for normalizing the vector composed of the environment elements $e$; $I()$ represents the method for normalizing the vector composed of the Agent collaborative network elements $a$; $U()$ is the representation method of the system effectiveness $u$; $\tau^c$ represents the scenario set finally generated by the algorithm, which contains $c$ scenarios, and the sum of the probabilities of all scenarios occurring is less than or equal to 1; $G_{Scenes}(S)$ represents generating the minimum number of scenarios from the solution set $S$; $X$ represents the sample space used to generate the corresponding elements in the parameter vector $\varepsilon_t$; $\to$ indicates that the $n$-dimensional vector $\varepsilon_t$ representing the state of each time - instant of each scenario $\tau^k$ in the scenario set $S$ is obtained by sampling from $n$ value distributions in the set $X$ respectively.

The system effectiveness $u_t$ at time period $t$ serves as the response variable of the system, which is used to reflect the indicator that researchers focus on, representing the overall state of the system at time period $t$. According to~\cite{yu2024beyond}, Value Entropy is used to measure the overall health degree of the system. Assuming that the total number of individuals in the system is $N_{total}$, based on the calculation method of value entropy, the system effectiveness $u_t$ at time period $t$ is defined as shown in the following formula:
\begin{equation}
    \label{eq:U}
    u_t = e^{1 - \left|\frac{-\sum_{j = 1}^{m} p_j \log_2 p_j - \log_2 \sqrt{N_{total}}}{\log_2 \sqrt{N_{total}}}\right|}
\end{equation}
\begin{equation}
    \label{eq:p}
    p_j =\frac{N_j}{N_{total}}
\end{equation}
Here, $m$ represents the number of niches divided in the service ecosystem;  $N_j$ is the number of individuals in niche $j$. $\log_2 \sqrt{N_{total}}$ is the optimal entropy of the system; $-\sum_{j = 1}^{m} p_j \log_2 p_j$ is the current entropy of the system.

The relative distance between the feature vector $\varepsilon_t$ of the generated scenario and the features of the original scenario is adopted to measure the degree $\delta$ to which the generated scenario $\tau$ preserves the characteristics of the initial scenario. Its formula is expressed as follows:
\begin{equation}
    \label{eq:Dvalue}
    \delta = \frac{\sum_{l = 1}^{L}|\tau_t^i - Es(t)|}{\sum_{l = 1}^{L}|Es(t)|} \times 100\%
\end{equation}
Here, $L$ is the dimension of the scenario parameter vector $\varepsilon$; $Es(t) (t = 1,2,\ldots,T)$ is the scenario composed of the expected values of historical scenarios in time $T$.

As systematically depicted in Fig.\ref{fig2}, base on tasks, domain data and knowledge, the scenario generator framework operates through three LLM-powered agents in a closed-loop architecture: The Planner Agent (PA) initiates the process by synthesizing system constraints through domain knowledge integration, followed by the Environment Agent (EA) generating extreme scenarios with long-tail distributions via adversarial semantic modeling. Concurrently, the Social Agent (SA) constructs heterogeneous agent populations by extracting social network backbones and behavioral strategy library \cite{zhou2024hierarchical}. These components collectively formulate multi-temporal system states $\varepsilon_t$, which feed into iterative experimental calibration through a feedback mechanism, ultimately enabling self-optimizing generation of scenario sets $S = \{\tau^1,\tau^2,\ldots,\tau^n\}$ for adaptive service ecosystem experimentation.

\subsection{The Environment Generation by EA}
EA aims to calculate the upper and lower limit boundaries of the environmental distribution based on the historical dataset, domain knowledge base, and adversarial prompt set. The detail is shown in Algorithm \ref{alg:EA}. The innovation of EA is mainly reflected in two aspects. 

1) In the scenario generation stage, by integrating the semantic understanding ability of LLM, the professional knowledge of the domain knowledge base, and adversarial prompts, it breaks through the limitation of traditional methods that only rely on the statistical features of historical data. It can generate more diverse and realistic scenarios, especially extreme scenarios with long-tail distributions. 

2) The scenario adversarial verification mechanism within the reinforcement learning framework is introduced to evaluate and screen the credibility of the generated scenarios, ensuring that the high-risk scenarios finally used for calculating the boundaries have high authenticity and reliability. This improves the accuracy and effectiveness of calculating the environmental distribution boundaries, providing a more scientific and comprehensive basis for demand environment modeling. 

\begin{algorithm}[ht]
\label{alg:EA}
\caption{Demand Environment Modeling and Extreme Environment Generation Algorithm by EA}
\KwIn{Historical dataset $D$, Domain knowledge base $K$, Adversarial prompt set $P$}
\KwOut{Upper and lower limit boundaries of the environmental distribution, $V_{max}$, $V_{min}$}
Initialize $C \gets \varnothing$, $S \gets \varnothing$, $R \gets \varnothing$, $C_{high_{risk}} \gets \varnothing$\;
\For{each $d$ in $D$}{
    Extract semantic features $c$ from $d$ using LLM and add $c$ to $C$\;
}
Extract event logic rule set $L$ from $K$ via knowledge distillation\;
\For{each $p$ in $P$}{
    Initialize $S_p \gets \varnothing$\;
    \For{each $c$ in $C$}{
        \For{each $l$ in $L$}{
            Generate scenario $s$ by combining $c$, $l$, and $p$ using LLM and add $s$ to $S_p$\;
        }
    }
    $S \gets S \cup S_p$\;
}
\For{each $s$ in $S$}{
    Set up an adversarial RL environment with $s$ as input, calculate credibility score $r_s$ and add it to $R$\;
}
Set thresholds $\theta_{high}$ and $\theta_{risk}$\;
\For{$i \gets 1$ \KwTo $|S|$}{
    \If{$R[i] \geq \theta_{high}$}{
        Calculate risk value $risk_i$ for $S[i]$\;
        \If{$risk_i \geq \theta_{risk}$}{
            $S_{high_{risk}} \gets S_{high_{risk}} \cup \{S[i]\}$\;
        }
    }
}
Extract data vector set $V$ from relevant variables in $S_{high_{risk}}$\;
Calculate lower and upper limit boundaries of environmental distribution from min and max values of variables in $V$\;
\Return The upper and lower limit boundaries of the environmental distribution as $V_{max}$, $V_{min}$\;
\end{algorithm}

\subsection{The Social Collaboration Generation by SA}
SA takes the social system data $D$ as input and aims to generate the backbone network $N$ of the relational structure. The detail is shown in Algorithm \ref{alg:SA}.  Its innovations are reflected in two aspects.

1) During the feature extraction phase, by leveraging the powerful semantic understanding and analysis capabilities of LLMs, it can deeply explore the latent characteristics of individuals and groups. Compared with traditional methods, it can obtain more abundant and accurate information, providing a solid foundation for subsequent relational network construction and behavior strategy generation.

2) In the process of relational network construction, it comprehensively considers the relationships among individuals, between individuals and groups, and among groups. By setting reasonable relationship strength thresholds, it accurately screens out meaningful relationships, making the constructed relational network more consistent with the complex structure of the actual social system.

\begin{algorithm}[h]
\label{alg:SA}
\caption{Social Relationship Structure Generation Algorithm by SA}
\KwIn{Social system data $D$}
\KwOut{Skeleton of the relationship network $N$}
Initialize $I \gets \varnothing$, $G \gets \varnothing$, $I_{features} \gets \varnothing$, $G_{features} \gets \varnothing$, $N \gets \varnothing$\;
\For{each $r$ in $D$}{
    \If{$r$ is individual}{
        Add $r$ to $I$\;
    } \ElseIf{$r$ is group}{
        Add $r$ to $G$\;
    }
}
\For{each $i$ in $I$}{
    Add $\text{LLM}(i)$ to $I_{features}$\;
}
\For{each $g$ in $G$}{
    Add $\text{LLM}(g)$ to $G_{features}$\;
}
\For{each $i$ in $I$}{
    \For{each $j$ in $I$ where $i \neq j$}{
        Calculate $s_{ij}$ based on features of $i$ and $j$; \If{$s_{ij}$ meets threshold $\mathcal{M}_{ij}$}{Add $(i, j, s_{ij})$ to $N$\;}
    }
    \For{each $g$ in $G$}{
        Calculate $s_{ig}$ based on features of $i$ and $g$; \If{$s_{ig}$ meets threshold $\mathcal{M}_{ig}$}{Add $(i, g, s_{ig})$ to $N$\;}
    }
}
\For{each $g_1$ in $G$}{
    \For{each $g_2$ in $G$ where $g_1 \neq g_2$}{
        Calculate $s_{g_1g_2}$ based on features of $g_1$ and $g_2$; \If{$s_{g_1g_2}$ meets threshold $\mathcal{M}_{g_1g_2}$}{Add $(g_1, g_2, s_{g_1g_2})$ to $N$\;}
    }
}
\Return $N$\;
\end{algorithm}

\subsection{The Rule Generation and Scheme Calibration by PA}
PA takes industry specification documents $I$ and expert experience $E$ as inputs, aiming to generate executable logical expressions $L$ and an optimized experiment scheme $P$. The detail is shown in Algorithm \ref{alg:PA}. The innovations of PA are mainly reflected in two aspects.

1) During the generation of system constraint conditions, by integrating the Retrieval-Augmented Generation (RAG) technology with LLM \cite{JAS-2024-0856}, PA combines industry specification documents and expert experience. This approach enables the full exploration and utilization of domain knowledge, resulting in system constraint conditions that are more in line with real-world scenarios. Compared with traditional methods, the generated constraint conditions are more comprehensive and accurate.

2) In the processes of rule compilation and experiment plan optimization, PA employs Program-aided Prompting technology to automatically compile natural language rules into executable logical expressions and achieves self-calibrating optimization of the experiment plan. This not only improves the efficiency and accuracy of rule compilation but also allows for dynamic adjustment of the experiment plan based on experimental results. Consequently, the finally generated executable logical expressions and experiment plan can better meet practical needs, providing a more intelligent and efficient method for the generation of system rules. 

\begin{algorithm}[h]
\label{alg:PA}
\caption{Rule Generation Algorithm by PA}
\KwIn{Industry specification documents $I$, Upper and lower limit boundaries of the environmental distribution $V_{max}$ and $V_{min}$, Set of key relationships for each year $N$ (including initial relationships $N_0$)}
\KwOut{Executable logical expressions $L$, Optimized experiment scheme $P$}
Initialize system constraint set $C \gets \varnothing$\;
\For{each document $d$ in $I$}{
    $constraint \gets \text{RAG}(d,\text{LLM})$\;
    Add $constraint$ to the system constraint set $C$\;
}
Initialize executable logical expression set $L \gets \varnothing$, Initialize the environmental scheme $Ev^{(0)}$\;
\While{not converged} {
    Compute gradient: $\nabla J = \left( \frac{\partial J}{\partial E\nu}, \frac{\partial J}{\partial N} \right)$ where $J = J(E\nu, N)$\;
    Update parameters: $E\nu^{(k+1)} \leftarrow E\nu^{(k)} - \eta \cdot \nabla J$\;
    \If{$\lVert\nabla J\rVert < \epsilon$ ($\epsilon$ is the preset convergence threshold)} {Break}
}
\For{each constraint $c$ in $C$}{
    $expression \gets \text{Program\_aided Prompting}(c, \text{LLM})$\; 
    Add $expression$ to the executable logical expression set $L$\;
}
Initialize experiment scheme $P \gets \text{InitialScheme}(L)$\;
\While{not converged}{
    results $\gets$ \text{Execute}(P)\;
    feedback $\gets$ \text{Analyze}(results)\;
    $P \gets \text{Adjust}(P, feedback, L)$\;
    \If{isConverged(feedback)}{
        converged $\gets$ true\;
    }
}
\Return {$L, P$};
\end{algorithm}

\section{Experimental Design and Result Analysis of Computational Experiments} \label{em}

Compared with the current transaction-based revenue-sharing governance strategy adopted by the World's Largest Public API Marketplace, \enquote{RapidAPI}, as of March 2025, the number of developers has reached over 4 million, the number of APIs has exceeded 40,000. The ProgrammableWeb ecosystem governance adopted a non-direct revenue strategy and was finally forced to be decommissioned, which highlights the importance of service ecosystem governance. Many researchers intend to conduct in-depth studies on its governance approach. However, it is difficult to formulate experimental plans and the experimental efficiency is low. Based on such requirements, we carry out extensive experiments to answer the following research questions regarding the scenario generator design method proposed in this paper:
\begin{itemize}
    \item \textbf{RQ1:} Can it take into account the extreme scenarios in the governance of the service ecosystem?
    \item \textbf{RQ2:} Can it effectively represent the collaborative characteristics of individuals within the service ecosystem?
    \item \textbf{RQ3:} Are the generated scenarios more accurate and efficient?
\end{itemize}

\begin{figure*}[htbp]
\centering
\includegraphics[width=0.99\textwidth]{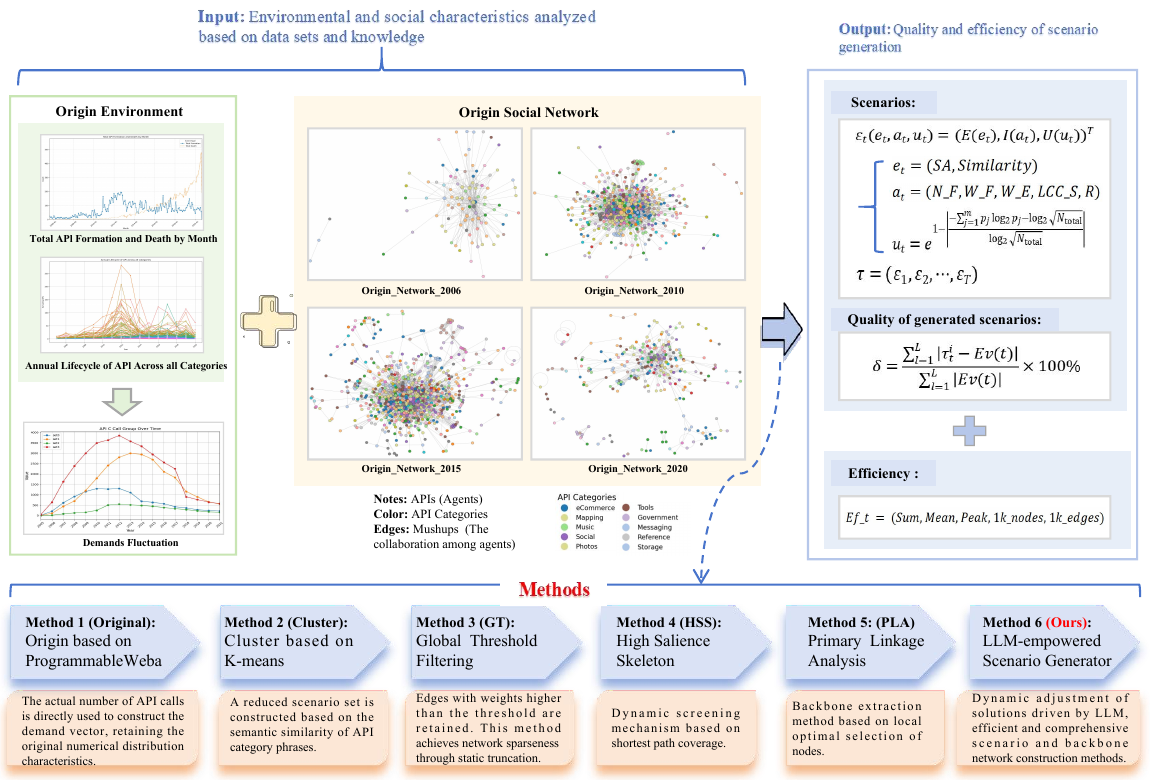}
\caption{The framework of computational experiments design. Using the analysis of environmental and social characteristics based on datasets and knowledge as input, six different experimental methods are employed for processing, and a scenario set along with evaluation indicators are outputted, including the assessment of the quality of generated scenarios and experimental efficiency.}
\label{fig3}
\end{figure*}

\subsection{Experimental Design}
ProgrammableWeb is an authoritative online platform that focuses on the API (Application Programming Interface) ecosystem. It was established in 2005. In the 2010s, with the explosion of the API economy (such as the open APIs of Twitter, Google Maps, etc.), ProgrammableWeb became an industry benchmark. As market competition intensified (e.g., RapidAPI was founded in 2014), it faced the problems of user loss and declining revenue, making it difficult to cover the maintenance costs. As a result, it was deactivated in 2022. Based on the ProgrammableWeb dataset, we designed experiments to analyze the evolution process of the ProgrammableWeb service ecosystem within its life cycle under a fluctuating social environment. 

As shown in Fig.\ref{fig3}, based on the analysis of the ProgrammableWeb dataset, social characteristics of environments and entities were extracted to form an initial distribution from raw data. Different experimental schemes were generated under distinct experimental design methods. Finally, quality and efficiency comparisons were conducted for scenarios generated by different schemes. In addition to \textbf{Method 1 (Original)} using raw data statistical analysis, we classified current research techniques in service ecosystem governance into four experimental design methods based on the summary of complex network Analysis and ABM in Section \ref{bk}:

\textbf{\textit{Experimental Methods:}} 
\begin{itemize}
    \item \textbf{Method 2: Cluster.} The K-means algorithm is used to partition the initial distribution into clusters. The cosine similarity between samples and cluster centers is calculated, and adaptive thresholding is then used to remove outliers and noise, producing the final experimental plan.
    \item \textbf{Method 3: Global Threshold Filtering (GT). } Through static truncation, edges with weights exceeding a predefined threshold are retained, enabling sparsification of samples under given environmental states and narrowing the distribution of generated scenarios.
    \item \textbf{Method 4: High Salience Skeleton (HSS).} Significant service dependency relationships are preserved based on statistical hypothesis testing. A single-source Dijkstra’s algorithm is applied to generate a shortest path tree for each node. The significance score of each edge is calculated according to path coverage frequency, and edges with scores above the threshold are retained.
    \item \textbf{Method 5: Primary Linkage Analysis (PLA).} For each element or node, the adjacent edge with the maximum weight is selected, and only bidirectional maximum-weight connections are retained to form an undirected network.
\end{itemize}

\textbf{\textit{Scenario Vector:}} Based on Formula (\ref{eq:S}), the scenario vector during the experiment is represented as follows:
\begin{equation}
    \label{eq:et}
    e_t = (SA, Similarity)
\end{equation}
Here, $SA$ (i.e., $Sum Activity$) is the total activity volume, which is the sum of times count APIs are called by Mashup, reflecting demand. APIs are divided into four categories by property. $Similarity$ monitors changes in market demand structure for each year according to the proportional distribution of the count APIs are called by Mashup.
\begin{equation}
    \label{eq:at}
    a_t=(NF, WF, WE, LCC\_S,Reachability)
\end{equation}
Here, $NF$ (i.e., $Node Fraction$) is the sub-graph to original graph node ratio, reflecting node retention integrity. $WF$ (i.e., $Weight Fraction$) is the ratio of retained edge to original network total weight, indicating core connection retention degree. $WE$ (i.e., $Weight Entropy$) is the edge weight probability distribution, measuring weight distribution diversity. $LCC\_S$ (i.e., $LCC Size$) is the largest connected component's node number, measuring network structure integrity. $Reachability$ is the proportion of node pairs with at least one path, showing network global connectivity and structural compactness. A value nearer 1 means a more compact network.

For API nodes of each year, logarithmic standardization is performed by comprehensively considering the characteristics of market demand and network structure. The total number of APIs is $N_{total}$, and the formula for calculating the individual effectiveness $I\_U_i$ based on Formula (\ref{eq:U}) is as follows:
\begin{equation}
    \label{eq:ui}
    I\_U_i=\alpha\cdot\ln\left(\frac{C_i}{C_{\max}} + 1\right)+\beta\cdot\ln\left(\frac{D_i}{D_{\max}} + 1\right)
\end{equation}
Here, $C_i$ is the total number of times the current API node is called by Mashup in the current year. $C_{\max}$ is the maximum number of times an API node is called by Mashup among all API nodes in the current year. $D_i$ is the network weighted degree centrality, with the formula $\sum_{j\in N(i)} w_{ij}$, where $w_{ij}$ represents the edge weight. 
$D_{\max}$ is the maximum weighted degree in the network in the current year. $\alpha$ is the weight coefficient for market demand, adjusting its contribution. $\beta$ is the weight coefficient for network influence, regulating the contribution of the collaborative network.

Sort the individual utility values in descending order and use the equal-width binning method to define $m$ niches ($m\approx\sqrt{N_{total}}$)\cite{yu2024beyond}. Based on Formula (\ref{eq:U}), the system effectiveness $u_t$ during time period $t$ is derived. 

\textbf{\textit{Evaluation Indicators:}}
The quality of generated scenarios is obtained based on Formula (\ref{eq:S}) and (\ref{eq:Dvalue}). The experimental efficiency is represented by the following indicators.
\begin{itemize}
    \item \textit{Sum(ks):} Total time of multiple experiments, measuring resource consumption in long-term or batch tasks.
    \item \textit{Mean(ks):} Average time per single experiment, reflecting processing efficiency in regular scenarios.
    \item \textit{Peak(ks):} Maximum single-experiment time during the process, evaluating method stability in the worst - case.
    \item \textit{Time per 1k nodes (/1kn) and per 1k edges (/1ke):} Quantify the sensitivity of different methods  to node and interaction scale. Lower values imply stronger large-scale network processing.
\end{itemize}

\subsection{Experimental Analysis}
Each experimental design method yields a final scheme. Executing the final scheme generates a corresponding scenario set. We randomly select one scenario for in - depth analysis to sequentially answer the three research questions. 
\begin{figure}[htbp]
\centering
\includegraphics[width=0.48\textwidth]{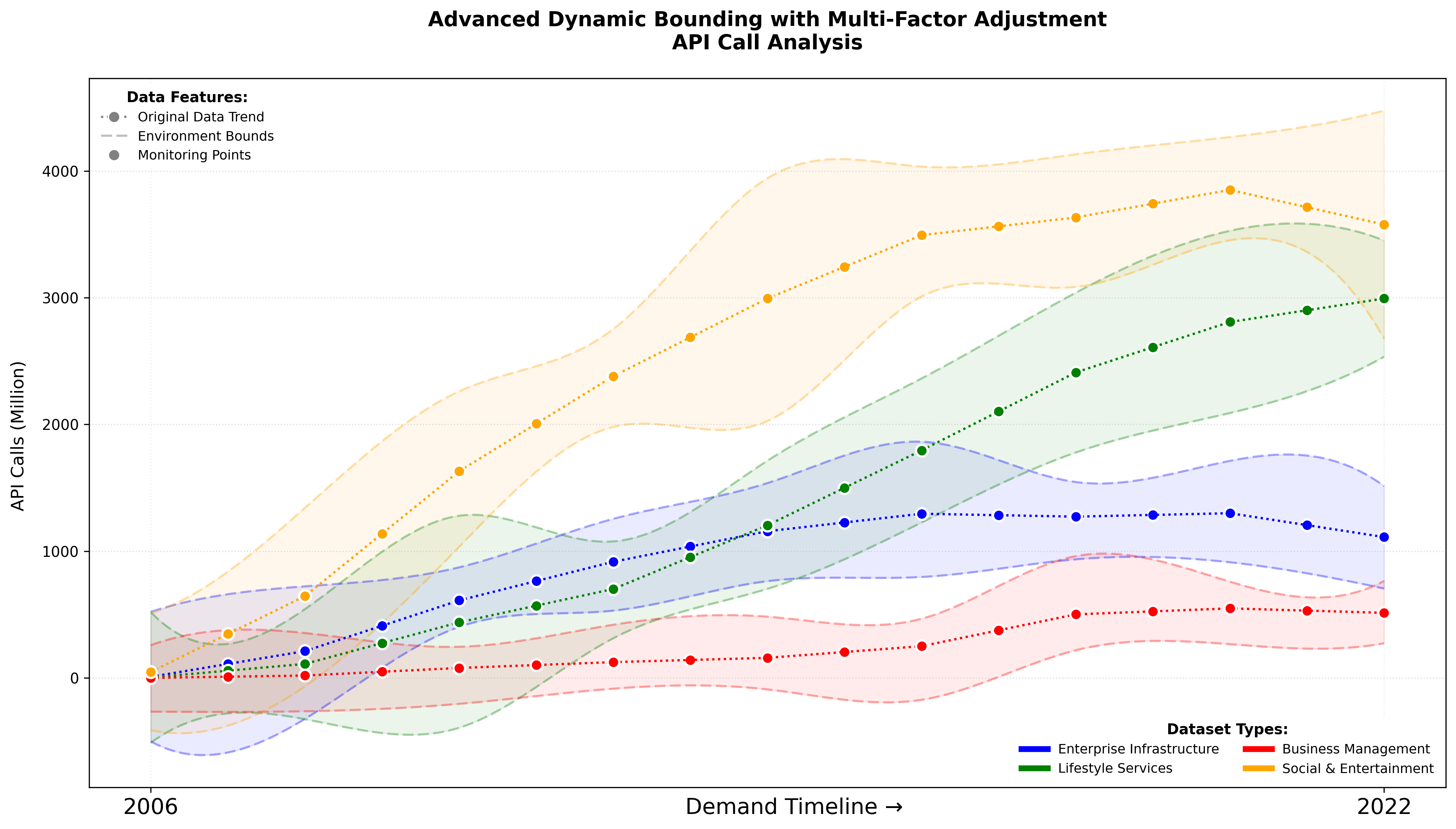}
\caption{Results of Extreme Environment Generation}
\label{fig4}
\end{figure}

\subsubsection{Analysis of the Generated Requirement Environment (RQ1)}
We collected the number of active APIs on ProgrammableWeb and RapidAPI from 2005 to 2022. The discrete API categories (such as eCommerce, Social) were mapped to a continuous semantic space using a pre-trained BERT model. Functionally similar categories were merged by an LLM agent, ultimately forming four major semantic communities. Finally, APIs were classified into four categories: Infrastructure, Lifestyle Services, Business Management, and Social Entertainment. For each of these four API sets, the number of times they were called by Mashups was counted to construct a demand time series, which was used to analyze changes in external market environment.

Based on this, the environment agent in the scenario generator fits the evolution trend of market demand. The LLM flexibly represents composite patterns such as growth, saturation, and oscillation by collecting online information and making inferences. By analyzing various data, the impact weights at different time scales are quantified. As shown in Fig.\ref{fig4}, our method determines the extreme boundary of environmental fluctuations at different times, and the environmental coverage has been significantly improved. 
\begin{figure*}[htbp]
\centering
\includegraphics[width=0.99\textwidth]{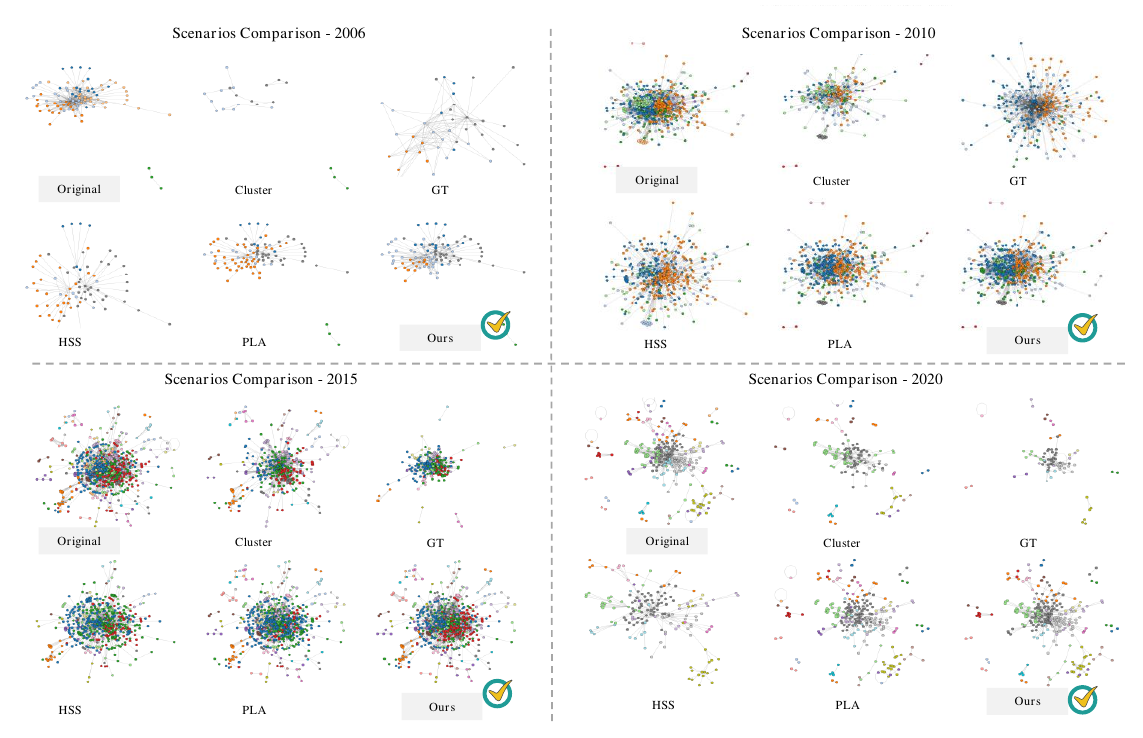}
\caption{The results of social collaborative relationship generation. It shows the comparison results generated by six methods, including Original, Cluster, GT, HSS, PLA and Ours, throughout the life course of  ProgrammableWeb's early (2006), mid-term (2010 and 2015) and late (2020) development. Among these methods, our method is most consistent with the original relational structure.}
\label{fig5}
\end{figure*}

\begin{table*}[htbp]
    \caption{Statistical Data on 8-dimensional Scenario Vector Indexes and Total Deviation of Each Method}
    \label{tb:eight}
    \centering
    \begin{tabularx}{\textwidth}{l *{9}{X}}
        \toprule
        Method & SA & Similarity & NF & WF & WE & LCC\_S & Reachability & VE (u) & Deviation($\delta$) \\
        \midrule
        Original & 0.9776 & 1 & 1 & 1 & 1 & 1 & 0.9712 & 0.8608 & --  \\
        Cluster & 0.3587 & 0.622 & 0.4643 & 0.226 & 0.8272 & 0.418 & 0.7886 & 0.9865 & 0.4329 \\
        GT & 0.8578 & 0.8179 & 0.3942 & 0.5397 & 0.7869 & 0.3894 & 0.9427 & 0.9093 & 0.2972 \\
        HSS & 0.92287 & 0.7753 & 0.7816 & 0.2214 & 0.7009 & 0.4214 & 0.3885 & 0.8114 & 0.3556 \\
        PLA & 0.9448 & 0.7987 & 1 & 0.2997 & 0.7087 & 0.674 & 0.4826 & 0.8049 & 0.2685 \\
        \hline
        Ours & 0.9817 & 0.9767 & 0.9978 & 0.6335 & 0.8973 & 1 & 0.9755 & 0.824 &\textbf{0.0783} \\
        \bottomrule
    \end{tabularx}
\end{table*}

\begin{table}[htbp]
    \caption{Statistics on Experimental Efficiency Times of Different Methods}
    \label{tb:time}
    \centering
    \begin{tabularx}{\columnwidth}{l *{5}{X}}
        \toprule
        Method & Sum(ks) & Mean(ks)  & Peak(ks) & /1kn(ks) & /1ke(ks) \\
        \midrule
        Original & 13.037 & 0.869 & 10.142 & 25.775 & 4.166 \\
        Cluster & 18.977 & 1.265 & 18.142 & 76.665 & 22.430 \\
        GT & 6.181 & 0.412 & 5.671 & 31.862 & 8.813 \\
        HSS & 5.709 & 0.381 & 5.314 & 14.329 & 14.884 \\  
        PLA & 5.937 & 0.396 & 5.393 & 11.739 & 12.701 \\
        \hline
        \textbf{Ours} & \textbf{3.171} & \textbf{0.211} & \textbf{1.585} & \textbf{6.279} & \textbf{1.907} \\
        \bottomrule
    \end{tabularx}
\end{table}

\subsubsection{Analysis of the Generated Social collaboration (RQ2)}
Social interaction characteristics within the ProgrammableWeb ecosystem are represented by an API-API co-occurrence network. Here, Mashup serves as the service composition unit, and API as the basic functional module. Weighted edges are constructed by counting the co-occurrence frequencies of APIs within the same Mashup, generating co-occurrence network graphs for the years 2005-2020. In the original state (Original), there are numerous API nodes with dense connections and complex co-occurrence relationships. The corresponding API types are diverse, conforming to the complexity of a real-world service ecosystem network. Based on the six methods above, an API-API co-occurrence network graph for each year is drawn. Nodes are colored according to the community division results of each network, as shown in Fig.\ref{fig5}.

To visually verify the ability of different methods to represent the core structure of the network, this paper selects the social association state diagrams of the initial state (2006), intermediate states (2010 and 2015), and the later state (2020) for comparative analysis from visual dimensions such as community consistency, key node retention, and color distribution fidelity. By analyzing the network evolution diagrams, it can be seen that the Cluster method disrupts the topology-demand coupling due to rigid semantic screening, leading to the simplification of community structure. GT and HSS, limited by static thresholds or path redundancy, sacrifice key connections for cross-category collaboration. The tree-like constraint of PLA causes over-screening and simplification, with the number of communities and color distribution deviating significantly from the benchmark network, poorly reflecting the mesh-like interactions in the real ecosystem. In contrast, Ours maintains a community structure and color distribution basically consistent with Original in all four stages, retains cross-community connections, and the color-mixing pattern is highly consistent with Original, being able to reflect the real-world collaboration pattern. 
\begin{figure}[htbp]
\centering
\includegraphics[width=0.499\textwidth]{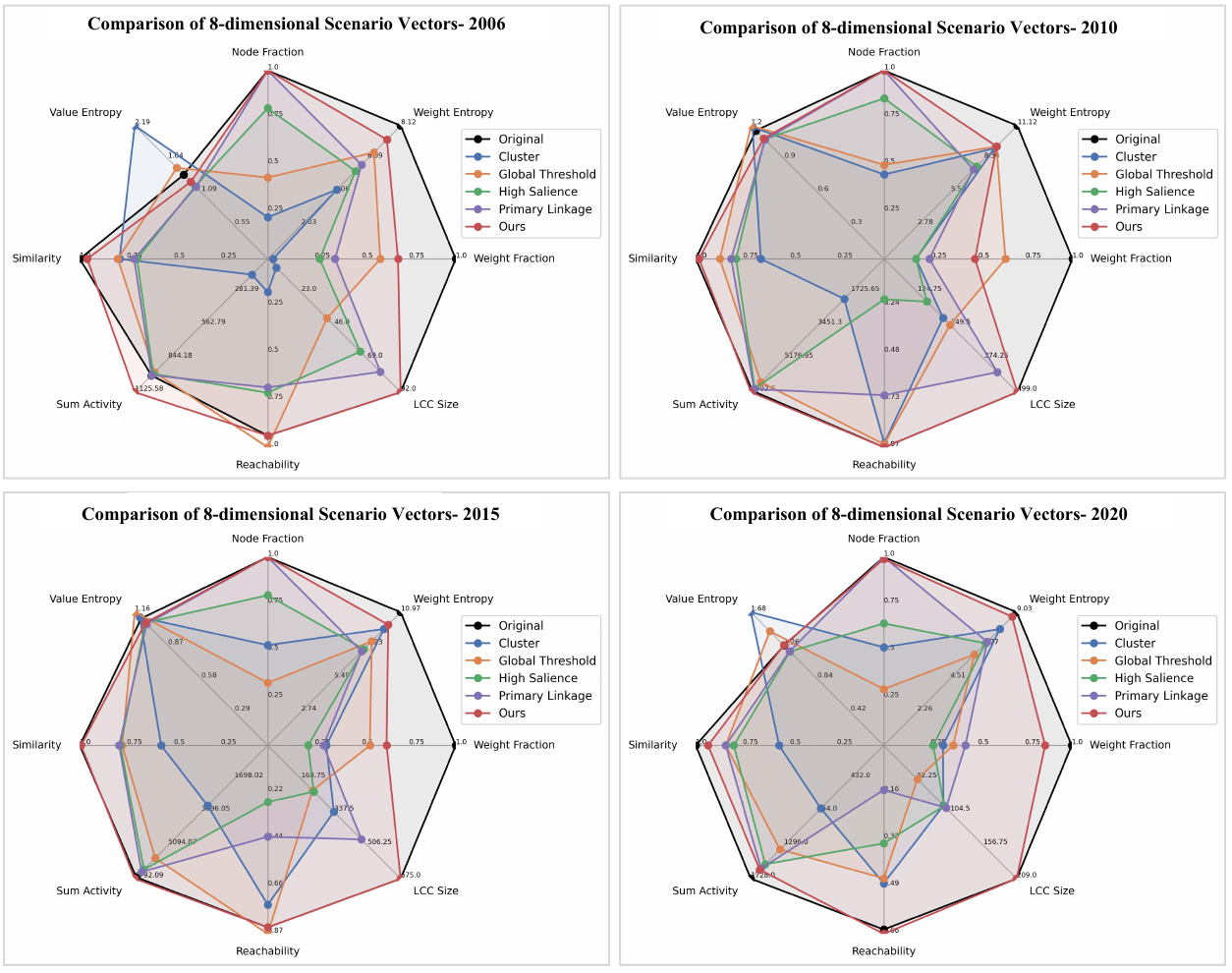}
\caption{The effect of scenario generator. It presents  the comparison of 8-dimensional scenario vectors for the scenarios generated by six methods: Original, Cluster, GT, HSS, PLA, and Ours. Among them, our method aligns most closely with the comprehensive features of the original scenario.}
\label{fig6}
\end{figure}

\subsubsection{Analysis of the Quality and Efficiency of Scenario Generator (RQ3)}
This paper comprehensively assesses the differences between the experimental design method of our proposed scenario generator and other methods through multi-dimensional analysis.

\textbf{Performance Evaluation.} Based on the calculation and comparison of the characteristic parameters of the 8-dimensional scenario vector, as shown in Fig.\ref{fig6}, taking the Original method as the benchmark to compare the scenario states in multiple time periods, the scenarios generated by our method show the best overall agreement with those of the Original method. In Table \ref{tb:eight}, the comprehensive difference degree ($\delta$) of our method in eight core indicators is only 0.0738, which represents a 72.5\% improvement in accuracy compared to the sub-optimal method PLA with a value of 0.2685. In the environmental dimension (i.e., SA, Similarity), the environmental features generated by our method are close to those of the Original method, with improvements of 14.5\% and 19.4\% respectively compared to the sub-optimal method GT. Moreover, other methods suffer from issues such as the loss of key APIs, sacrifice of long - tail requirements, or structural deviation. Regarding social relationship generation (i.e., NF, WF, WE, LCC\_S, Reachability), our method is close to the original network in terms of network integrity, global reachability, and weight distribution diversity. In contrast, methods like GT, Cluster, HSS, and PLA exhibit varying degrees of adaptability defects or connectivity losses. In terms of system health (i.e., VE), the deviation of our method from the Original network is significantly better than other methods.

\textbf{Experimental Efficiency.} Calculated based on Table \ref{tb:time}, the total time for multiple rounds of experiments of our method is 3.171 thousand seconds. This is 75.7\% lower than the 13.037 thousand seconds of the benchmark Original method and 44.5\% more optimized than the 5.709 thousand seconds of the sub-optimal method HSS. Our method outperforms in core indicators such as time per thousand nodes, time per thousand edges, and peak time. It also demonstrates excellent stability, with the average time being significantly better than that of the comparison methods. Additionally, the growth rate of the time per thousand nodes/edges of our method is lower than the network scale growth rate, indicating its potential to handle large-scale data and providing methodological support for the research on the governance of service ecosystems.

In conclusion, through dynamic weight adjustment and spatio-temporal coupling modeling, our method effectively resolves the performance contradictions of existing methods. It achieves coordinated optimization in environmental distribution, integrity of social relationship structure, and improved experimental efficiency greatly, thus providing a new paradigm for conducting experiments in the research on the governance of service ecosystems. 

\section{Conclusion} \label{cl}

In summary, this study proposes an LLM-empowered agents collaboration framework for generating governance-referenced scenarios in dynamic service ecosystems. The approach coordinates three specialized agents: the Environment Agent (EA) simulates extreme social environments, the Social Agent (SA) generates features and collaborative networks, and the Planner Agent (PA) couples task-role relationships and optimizes task solutions. Through autonomous experimental regulation, the framework quantifies qualitative features and produces high-quality, computationally efficient scenario generation schemes. Experiments on the ProgrammableWeb dataset demonstrate that the method generates more accurate scenarios, offering a novel and effective approach for constructing experimental systems tailored to service ecosystem governance.

Despite its strengths, the current framework has limitations, including potential semantic biases inherent in LLMs and scalability constraints when handling large-scale, complex scenarios. Future research should focus on integrating reinforcement learning for adaptive exploration, developing lightweight verification modules to enhance reliability, and incorporating causal frameworks to better address intricate social interdependencies. These advancements will further strengthen the framework, providing robust governance tools for evolving service ecosystems and enabling more effective decision-making in dynamic social environments.

\section*{Acknowledgment}
\addcontentsline{toc}{section}{ACKNOWLEDGMENT}
This work has been supported in part by the Natural Science Foundation of China (No.92367202), National Natural Science Foundation of China (No.62472306, No.62441221, No.62206116), Tianjin University's 2024 Special Project on Disciplinary Development (No.XKJS-2024-5-9), Tianjin University Talent Innovation Reward Program for Literature \& Science Graduate Student (C1-2022-010), and Henan Province Key Research and Development Program (No.251111210500), Shandong Taishan Industry Leading Talent Project (NO.tscx202211010).

\bibliographystyle{IEEEtran}
\bibliography{ref}

\end{document}